\pdfoutput=1
\documentclass[11pt]{article}

\usepackage[]{EMNLP2022}

\usepackage{microtype}
\usepackage{graphicx}
\usepackage{subcaption}
\usepackage{booktabs} % for professional tables
\usepackage{tikz} % for overview figures
\usepackage{tikz-layers} % for overview figures
\usepackage{xcolor} % for colors
\usepackage{amsmath} % for math symbols
\usepackage{amssymb} % for math symbols
\usepackage{wasysym} % for math symbols
\usepackage{array}
\usepackage{times}
\usepackage{latexsym}

\usepackage[frozencache,cachedir=.]{minted}
\usepackage{comment}
\usepackage{rotating}

\usepackage[T1]{fontenc}
\usepackage[utf8]{inputenc}

\usepackage{fancyhdr}
\title{SMART: Sentences as Basic Units for Text Evaluation}

\author{
Reinald Kim Amplayo, Peter J. Liu, Yao Zhao, Shashi Narayan\\ 
Google Research \\ 
\texttt{\small \{reinald, peterjliu, yaozhaoyz, shashinarayan\}@google.com}
}

\begin{document}
\maketitle

\begin{abstract}
Widely used evaluation metrics for text generation either do not work well with longer texts or fail to evaluate all aspects of text quality. In this paper, we introduce 
% extend ROUGE into 
a new metric called SMART to mitigate such limitations. Specifically, We treat sentences as basic units of matching instead of tokens, and use a sentence matching function to \textit{soft}-match candidate and reference sentences. Candidate sentences are also compared to sentences in the source documents to allow grounding (e.g., factuality) evaluation. Our results show that system-level correlations of our proposed metric with a model-based matching function outperforms all competing metrics on the SummEval summarization meta-evaluation dataset, while the same metric with a string-based matching function is competitive with current model-based metrics. The latter does not use any neural model, which is useful during model development phases where resources can be limited and fast evaluation is required. Finally, we also conducted extensive analyses showing that our proposed metrics work well with longer  summaries and are less biased towards specific models.
%, and selects better model checkpoints than ROUGE. 
\end{abstract}
\section{Introduction}

\begin{figure}[t]
    \centering
    \includegraphics[width=\columnwidth]{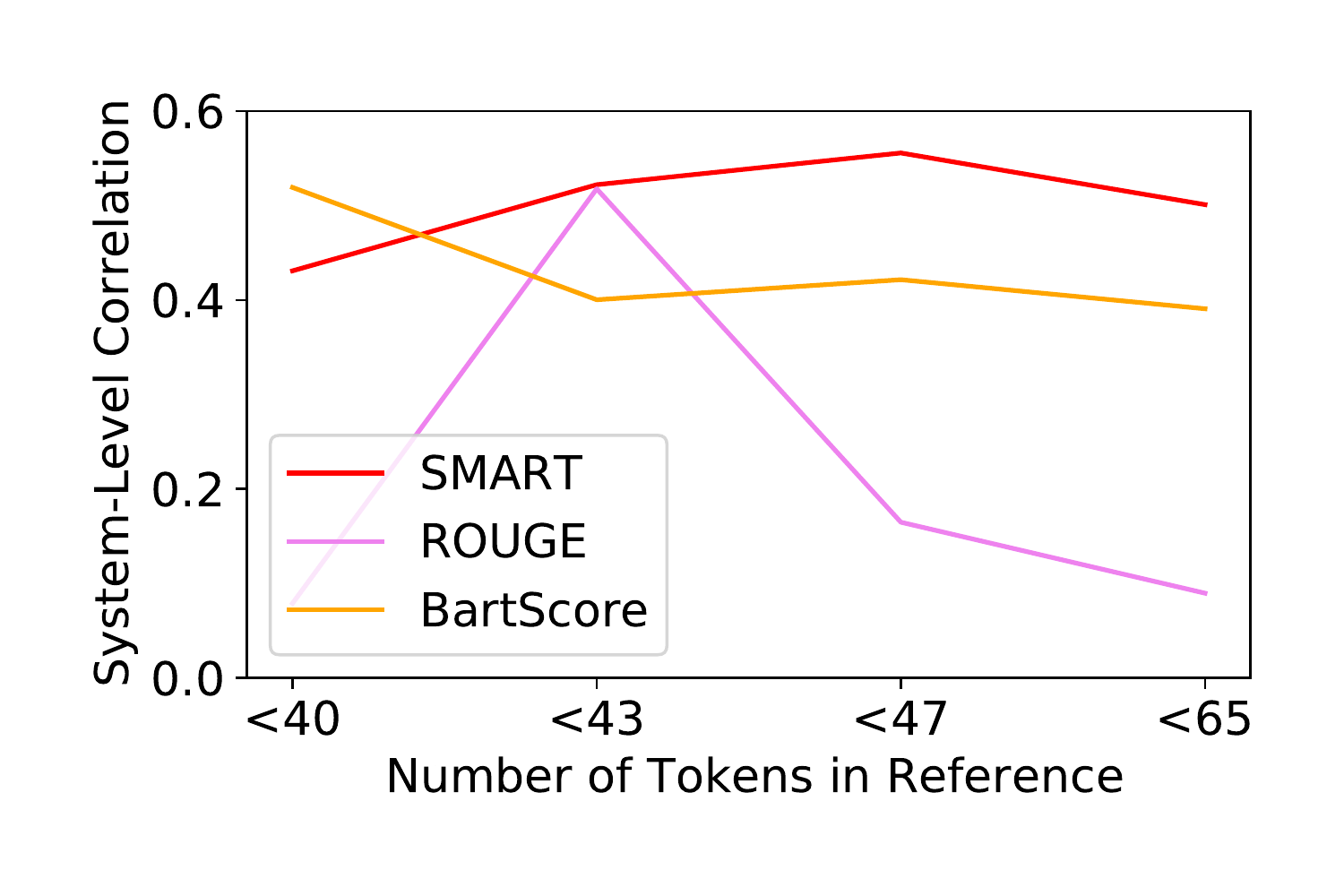}
    \caption{Kendall tau system-level correlations of ROUGE and SMART averaged over four dimensions of summary quality as the number of tokens increases. Summaries are from CNN/DM \cite{hermann2015teaching} and human annotations are from SummEval \cite{fabbri-etal-2021-summeval}. Each bucket in the x-axis contains equal number of data points. More details in Section \ref{sec:analysis}.%.\sn{Could you compress this figure by 50\% in height? Could we add few others string-based and model-based metrics?}
    }
    \label{fig:intro}
\end{figure}

One major obstacle in the progress of text generation tasks (e.g., document summarization, long-form question answering, data-to-text generation, etc.) is automatic evaluation. 
Traditionally, automatic metrics that rely on discrete token-level matching such as ROUGE \cite{lin-2004-rouge} and BLEU \cite{papineni-etal-2002-bleu} have been utilized to check whether system outputs are of high quality across four dimensions \cite{kryscinski-etal-2019-neural,yuan2021bartscore}: coherence, factuality, fluency, and informativeness.
These metrics do not correlate well with human judgments on all four dimensions of text quality \cite{fabbri-etal-2021-summeval}. Because of this, the evaluation is usually coupled with human elicitation studies that ask humans to rate texts. These studies can be expensive and nearly impossible to reproduce.

More recently, pretrained language models are leveraged for automatically evaluating system-generated texts \cite{Zhang*2020BERTScore:,sellam-etal-2020-bleurt,yuan2021bartscore}, which have shown improvements on correlation with human judgments. 
Nevertheless, both ROUGE and LM-based metrics have three major drawbacks.
Firstly, these metrics are not good at evaluating long and multi-sentence texts.
Figure \ref{fig:intro} illustrates system-level rank correlations of ROUGE in different text lengths, which shows that after a certain length, ROUGE drastically decreases its evaluative power.
By design, ROUGE is also not robust to evaluating possibly shuffled information in long outputs, hurting its performance on evaluating coherence.
On the other hand, LM-based metrics such as the state-of-the-art BARTScore \cite{yuan2021bartscore}, are constrained to the length limitation of the pretrained LM used, thus they are not able to evaluate outputs longer than this limit.
Secondly, most of these metrics only use reference texts during evaluation. This restricts the capability of the metrics from evaluating dimensions of text quality that requires grounding to the source.
\citet{yuan2021bartscore} suggested to use the source document during evaluation, however their evaluation is still limited to short documents because of length limitations in LMs.
Finally, LM-based metrics can be very slow, which hinders usage in the model development stages, where quick evaluation is necessary.
As such, most practitioners use ROUGE in these stages, which can lead to suboptimal modeling.

In this paper, we propose an automatic metric called SMART (\textbf{S}entence \textbf{MA}tching for \textbf{R}ating \textbf{T}ext).
% , which can be seen as an extension of ROUGE. 
To support long and multi-sentence texts, we treat sentences as basic units of matching instead of tokens.
Since sentences most likely do not have exact matches, we use a soft-matching function that returns a matching score between 0 and 1, given a pair of sentences. 
We experiment with several string-based and powerful model-based soft-matching functions with SMART (see \S\ref{sec:matching_fn}).
% This allows us to leverage highly correlated sentence-level machine translation metrics such as BLEURT \cite{sellam-etal-2020-bleurt}.
Moreover, to allow grounded evaluation, we also include the source in the calculation of the metric.
Similar to ROUGE, we introduce multiple SMART versions using sentence n-gram overlap and longest common subsequence. We show that SMART with BLEURT \cite{sellam-etal-2020-bleurt} as a soft-matching function outperforms all the approaches we compared against on all four dimensions of quality 
% (coherence, factuality, fluency, and informativeness) 
in the SummEval dataset \cite{fabbri-etal-2019-multi}.
Moreover, a faster variant of SMART, which does not use any neural model for text matching, shows competitive correlations with human judgments. 
Finally, our extensive analyses show that SMART works better with longer summaries and is less biased towards specific models.
%, and can select better model checkpoints than ROUGE.
We will release and maintain a toolkit containing scripts to easily run evaluation using SMART.
\section{Related Work}

\paragraph{String-based Text Evaluation}

Evaluation in conditional generation tasks such as machine translation and document summarization is a long-standing problem. Traditionally, evaluation involves human elicitation studies that score texts based on different metrics of quality, such as adequacy, fidelity, and fluency in machine translation \cite{hovy1999toward}, and coherence, conciseness, fluency, readability, and content relevance in summarization \cite{mani2001automatic}. 
Automatic metrics based on token n-gram matching have been developed to replace these expensive and time-consuming studies, in which ROUGE \cite{lin-2004-rouge} and BLEU \cite{papineni-etal-2002-bleu} are most widely used in summarization and translation, respectively.
Several extensions to token n-gram matching have been proposed, such as using paraphrases, synonyms \cite{lavie-agarwal-2007-meteor}, and word embeddings \cite{ng-abrecht-2015-better} to handle cases that are semantically equivalent, and downweighting common n-grams to focus more on salient ones \cite{vedantam2015cider}. \citet{popovic-2015-chrf} instead use character-level n-gram matching to also match words that are conjugated differently and support morphologically rich languages. 

\paragraph{Model-based Text Evaluation}

With the introduction and success of pretrained language models such as BERT \cite{devlin-etal-2019-bert} and BART \cite{lewis-etal-2020-bart}, evaluation metrics that leverage them have been proposed. BERTScore \cite{Zhang*2020BERTScore:} leverages contextualized token embeddings from BERT and obtains pairwise matching of tokens from reference and system summaries. MoverScore \cite{zhao-etal-2019-moverscore} extends BERTScore by instead having many-to-one soft alignments using Word Mover's Distance (WMD; \citealp{kusner2015from}). BLEURT \cite{sellam-etal-2020-bleurt} fine-tunes BERT to predict human scores with large-scale synthetic training data. BARTScore \cite{yuan2021bartscore} uses BART and treats evaluation as a text generation problem, using likelihood of predicting the system summary given the source document or the reference summary.
\citet{clark-etal-2019-sentence} and \citet{zhao-etal-2019-moverscore} also explored sentence-level matching with WMD using (contextualized) sentence embeddings, however they show no concrete improvements over other model-based metrics \cite{fabbri-etal-2021-summeval}.
In contrast, we show that our metric which uses sentence-level matching correlates better with human judgments than all competing models.

% factuality evaluation
Factuality in summarization \cite{falke-etal-2019-ranking,maynez-etal-2020-faithfulness} is usually evaluated separately since most automatic metrics are focused on informativeness and do not include the source document in the metric calculation. Factuality-specific metrics can be divided into three approaches: natural language inference (NLI) based approaches \cite{falke-etal-2019-ranking,maynez-etal-2020-faithfulness,laban-etal-2022-summac}, where a summary is considered factual if all its facts are entailed by the source document, model-based approaches \cite{kryscinski-etal-2020-evaluating,deng-etal-2021-compression}, where a model is trained to detect factual errors in the summary, and question answering (QA) based approaches \cite{durmus-etal-2020-feqa,wang-etal-2020-asking,honovich-etal-2021-q2}, where questions generated in a factual summary should be answerable using the source.
While we also compare correlations of automatic metrics with human judgments on factuality, the goal of our work is to find holistic metrics for evaluation that can also effectively evaluate other dimensions of text quality. 
Our results show that other dimensions such as coherence, fluency, and informativeness also benefit in the use of the source documents in the metric.
Finally, our results also show that our metric is at least comparable with both NLI- and QA-based factuality specific metrics \cite{honovich-etal-2021-q2,honovich2022true} in evaluating factuality, while outperforming them in evaluating other dimensions of quality.
While there are meta-evaluation datasets that focus mainly on factual consistency evaluation \cite{falke-etal-2019-ranking,wang-etal-2020-asking,honovich2022true}, we leave exploration of our metrics on those datasets as future work. 
\section{Problem Definition}

We use document summarization -- the task of generating concise and accurate summaries of input documents \cite{mani2001automatic} -- to explain our metric, but the proposed metric can be easily adapted to other text generation tasks.

Let  $\mathcal{S}$ be a list of source documents, $\mathcal{C}$ be a list of summaries generated for $\mathcal{S}$ by a candidate system, and $\mathcal{R}$ be a list of reference summaries produced by human annotators for $\mathcal{S}$. Note that $\mathcal{R}$ can be a nested list, i.e., for each example, there can be multiple references. 
Moreover, let $\mathcal{Q}$ be a list of dimensions of summary quality, and
let $\mathcal{H}_q$ be a list of human-annotated scores for $\mathcal{C}$ in terms of a certain summary quality $q$.
For each summary quality $q \in \mathcal{Q}$, the problem is to devise an evaluation metric $f_q(\mathcal{S},\mathcal{R},\mathcal{C})$ that outputs a list of scores that correlates well with $\mathcal{H}_q$.
Note that, unlike most of previous work on summarization evaluation \cite{lin-2004-rouge,clark-etal-2019-sentence,bhandari-etal-2020-evaluating,fabbri-etal-2021-summeval}, we also take into account source documents $\mathcal{S}$ when calculating metric $f_q(\cdot)$. This ensures that the metric can evaluate quality dimensions that require looking at the source.

We define the list of summary quality $\mathcal{Q}$ as the following four dimensions of summary quality, defined as follows (based on definitions in \citealp{fabbri-etal-2021-summeval} and \citealp{yuan2021bartscore}):

\begin{itemize}
    \item \textbf{Coherence}: The summary should be well-structured and well-organized. The summary should not just be a heap of related information, but should build from sentence to sentence to a coherent body of information about a topic.
    \item \textbf{Factuality}: The summary should only contain statements that are entailed by the source document. The summary should not contain hallucinated facts that either do not exist in the source or contradict facts from the source.
    \item \textbf{Fluency}: The summary should have no formatting problems, capitalization errors or obviously ungrammatical sentences (e.g., fragments, missing components) that make the text difficult to read.
    \item \textbf{Informativeness}: The summary should include only important information from the source document. The summary should not include redundant information and information that are considered excessive and non-salient.
\end{itemize}

\section{Sentence Matching for Rating Text}

We now describe our proposed metric, SMART (\textbf{S}entence \textbf{MA}tching for \textbf{R}ating \textbf{T}ext), which 
%is an extension of ROUGE with
has two key ideas. 
Firstly, we treat sentences as basic units of matching between system and reference summaries, instead of tokens. 
At sentence-level, exactly matching sentences are most likely non-existent (in datasets with abstractive reference summaries), thus we instead utilize soft-matching functions to compare sentences. Similar to ROUGE, we present two types of SMART: n-gram overlap (SMART-N) and longest common subsequence (SMART-L).
Secondly, SMART allows to compare the candidate system summary with \textit{both} the reference summary and the source document. This is particularly important when evaluating dimensions of summary quality that rely on the source document such as factuality.

\subsection{SMART-N}
\label{sec:sent-rn}

In order for SMART to work, summaries should be split into sentences. Let $R=[r_i]$ and $C=[c_j]$ be the sequence of sentences of the reference and the candidate system summary. 
SMART-N finds pairs of sentence n-grams in $R$ and $C$ that have the highest matching scores given by a sentence matching function that returns a score between 0 and 1 (detailed description in Section \ref{sec:matching_fn}).

Formally, given $N$ as the number of sentences in the sentence n-gram,
SMART-N can be calculated as follows:
\begin{align}
    {prec}' &= \sum\nolimits_{j=1}^{|C|-N+1} \text{max}_{r_i \in R; i \leq |R|-N+1} \nonumber\\
    &\quad \left[ \sum\nolimits_{n=0}^{N-1} \texttt{match} (c_{j+n}, r_{i+n}) / N \right] \\
    {prec} &= {prec}' / (|C| - N + 1) \\
    rec' &= \sum\nolimits_{i=1}^{|\mathcal{R}|-N+1} \text{max}_{c_j \in C; j \leq |C|-N+1} \nonumber\\
    &\quad \left[ \sum\nolimits_{n=0}^{N-1} \texttt{match} (r_{i+n}, c_{j+n}) / N \right] \\
    {rec} &= {rec}' / (|R| - N + 1) \\
    f &= 2 * prec * rec / (prec + rec)
\end{align}
where $\texttt{match}(\cdot, \cdot)$ is the sentence matching function, $prec$, $rec$, and $f$ are precision, recall, and f-measure, respectively. 
Note that unlike ROUGE, the numerators of precision and recall are different due to the use of a soft-matching function, thus they are calculated separately.
In our experiments, we used SMART-1 and SMART-2\footnote{
In special cases where either the candidate or the reference is a single-sentence summary, a normal implementation of SMART-2 would return zero since one of the summaries would have zero sentence bigrams. To mitigate this issue, we pad summaries with a blank sentence on both sides when calculating SMART-2. This ensures that we get a non-zero score for single-sentence summaries. In fact, SMART-2 reduces to SMART-1 in this case.
}, but SMART-N can be easily extended to work with larger $N$s.

\subsection{SMART-L}
\label{sec:sent-rl}

SMART-L is essentially the Longest Common Subsequence (LCS) of sentences in the reference and the candidate system summary. However, the original LCS algorithm requires an exact match to work.
We instead use a \textit{soft} version of LCS, where the task is defined as: 
Given two sequences $X=[x_i]$ and $Y=[y_j]$ and a matching function $\texttt{match}(x_i, y_j)$, find two \textit{soft}-subsequences $x_{i_1},...,x_{i_l}$ and $y_{j_1},...,y_{j_l}$ of length $l$ with $i_{k-1} \leq i_k \leq i_{k+1}$ and $j_{k-1} \leq j_k \leq j_{k+1}$, maximizing the sum $\sum\nolimits_{k=1}^l \texttt{match}(x_{i_k}, y_{j_k})$.

Unlike normal subsequences, soft-subsequences allow repetition of sentences as long as they do not go back to previous sentences (hence the use of $\leq$ operator). This relaxation helps in cases where the meaning of a sentence on one side spans over multiple sentences on the other side.
Furthermore, Soft-LCS is similar but different from a simple sequence alignment problem since the weight of the match depends on \textit{both} the positions of the items in the sequence and the items themselves. It is a less-restricted version of the Heaviest Common Subsequence (HCS; \citealp{jacobson1992heaviest}) since the matching function is relaxed to allow the use of a soft match (which is essentially an exact mismatch) in the subsequence. 

It turns out that Soft-LCS can be solved using a dynamic programming algorithm similar to that of LCS, which is illustrated as a pseudocode in Figure \ref{fig:lcs}. The main difference is that since we do not require an exact match, we always take the maximum among three cases: (1) choosing to soft-match $x_i$ and $y_j$, (2) choosing to soft-match $x_i$ and $y_{j-1}$, and (3) choosing to skip $x_i$.

\BeforeBeginEnvironment{minted}{\medskip}
\AfterEndEnvironment{minted}{\vspace{-2mm}}
\begin{figure}
    \begin{minted}
[
frame=lines,
framesep=2mm,
baselinestretch=1.0,
fontsize=\small,
]
{python}
def soft_lcs(X, Y):
  lcs = [[0] * (len(Y)+1)] * (len(X)+1)
  for i in range(len(X)+1):
    for j in range(len(Y)+1):
      if i != 0 and j != 0:
        m = match(X[i], Y[j])
        lcs[i][j] = max(lcs[i-1][j-1]+m,
                        lcs[i-1][j]+m,
                        lcs[i][j-1]) 
  return lcs[-1][-1]
\end{minted}
    \caption{Python pseudocode of the soft version of Longest Common Subsequence (Soft-LCS) given two sets of summary sentences X and Y.}
    \label{fig:lcs}
\end{figure}

Given the Soft-LCS function, we can then calculate SMART-L as follows:
\begin{align}
    prec' &= \texttt{soft-lcs}(C, R) \\
    prec &= prec' / |C| \\
    rec' &= \texttt{soft-lcs}(R, C) \\
    rec &= rec' / |R| \\
    f &= 2*prec*rec / (prec+rec)
\end{align}

\subsection{Comparing with Source}

Some dimensions of summary quality require access to source to be effectively evaluated. To cover those dimensions, SMART also compares the candidate system summary with the source, in addition to comparison with the reference summary. Let $S=[s_k]$ be the sequence of sentences of the source document. SMART that uses both source and reference is calculated as follows. We first calculate two SMART scores that (1) compares candidate system summary $C$ with reference summary $R$, and (2) compares $C$ with source document $S$. Then, we aggregate the scores by taking their maximum. For example, \texttt{SMART-N}$(S, R, C)$ is calculated as:
\begin{align}
    & \texttt{SMART-N}(S, R, C) = \text{max}(  \nonumber\\ 
    & \quad \texttt{SMART-N}(S, C), \texttt{SMART-N}(R, C) ) \label{eq:smart_with_src_ref}
\end{align}

\subsection{Multiple References}

Finally, when there are multiple reference summaries, we calculate SMART for each reference, and aggregate them by taking their maximum, as also commonly done in previous work \cite{fabbri-etal-2021-summeval}. This is intuitive since the candidate system summary only needs to match with at least one of the reference.

\subsection{Shorter Acronym}

We use the following template to describe SMART variants in a space-efficient manner:
\begin{quote}
    \centering
    \texttt{S[1|2|L]-m}
\end{quote}
where \texttt{m} is the sentence matching function of choice. For example, SMART-1 with a BLEURT \cite{sellam-etal-2020-bleurt} matching function can be shortened into S1-BLEURT.

%\subsection{Combining Source and Reference for Text Summarization Evaluation}

%The metrics introduced so far assume that we only have access to the reference summary (i.e., evaluation is done through text matching). However, when evaluating certain dimensions of quality such as coherence and factuality, including the source document during evaluation may be necessary. We introduce five different ways to use both/either the source document and/or the reference summary for text summarization evaluation: (1) \textbf{ref-only} metrics disregard the source entirely; (2) \textbf{src-only} metrics disregard the reference and use the source as $X$ instead of the reference in Sections \ref{sec:sent-rn} and \ref{sec:sent-rl}; (3-4) calculating both ref-only and src-only metric and combining them by taking the \textbf{average} and the \textbf{maximum}; and (5) \textbf{f1(src-p, ref-r)}: getting the f1-measure using the source document to calculate precision and the reference summary to calculate recall. The motivation behind the last alternative is that precision against the source document would check that all information in the candidate system summary is grounded from the source while recall against the reference summary would check that all information are considered informative.
\section{Sentence Matching Functions}
\label{sec:matching_fn}

The scores we get from SMART depend on the sentence matching function $\texttt{match}$. We investigate six different sentence matching functions widely used in both machine translation and document summarization literature. Specifically, we compare three string-based and three model-based matching functions. The former do not rely on accelerators (GPUs/TPUs) while the latter leverage pretrained neural models. Thus, we expect string-based matching functions to be inferior, but they are good alternatives for faster evaluation or when accelerators are not available.

\paragraph{ROUGE \cite{lin-2004-rouge}} A popular document summarization evaluation metric, it measures the number of overlapping textual units. As with most summarization work, we explored three types of textual units: unigrams (ROUGE-1), bigrams (ROUGE-2), and longest common subsequence (ROUGE-L).

\paragraph{BLEU \cite{papineni-etal-2002-bleu}} A popular machine translation evaluation metric, it is a precision-focused metric that calculates n-gram overlap between two texts and also includes a brevity penalty.

\paragraph{CHRF \cite{popovic-2015-chrf}} Another machine translation evaluation metric, it calculates character-based n-gram overlap between system and reference sentences. Unlike ROUGE and BLEU which operate at the token level, CHRF is more effective especially in morphologically-rich languages as it does not require any tokenization, lemmatization, and stemming \cite{mathur-etal-2020-results,kocmi-etal-2021-ship}.

\paragraph{BERTScore \cite{Zhang*2020BERTScore:}} A model-based metric that leverages contextualized token embeddings from BERT(-like models). It computes similarity scores by aligning tokens from reference and candidate summaries, and token alignments are computed greedily to maximize cosine similarity.
    
\paragraph{T5-ANLI \cite{honovich2022true}} Another model-based metric mainly used to evaluate factuality, which uses T5 \cite{raffel2020exploring} fine-tuned on the ANLI dataset \cite{nie-etal-2020-adversarial} to produce a score between 0 (not entailed) and 1 (entailed) given a premise and a hypothesis. We use the source/reference as premise and the candidate summary as hypothesis.

\paragraph{BLEURT \cite{sellam-etal-2020-bleurt}} A supervised model-based metric that uses BERT that is trained to predict human judgment scores using a small-scale dataset. To make it more robust, the model is first pretrained with a large-scale synthetic dataset. Moreover, it is optimized using several objectives including ROUGE, BLEU, BERTScore, and entailment. BLEURT has been shown to be effective in evaluating sentence match in machine translation, thus we expect it to be the better matching function among all matching functions.

One advantage of SMART is that it is easily extensible by changing the matching functions to better ones. This means that a more domain-specific matching function can be used for evaluation towards specific domains, or a better-performing sentence matching metric can be used to improve overall evaluation.
\section{Experiments and Results}

\subsection{Experimental Setting}

\paragraph{Dataset and Evaluation}
We conducted experiments on the SummEval dataset \cite{fabbri-etal-2021-summeval}, a document summarization meta-evaluation suite consisting of summaries from the CNN/DM dataset \cite{hermann2015teaching}.
Annotation is done in two stages and using experts to ensure high quality and high agreement across annotators. There are 1600 data points (16 systems $\times$ 100 examples) in total, each of which includes a score between 1 to 5 for each of the four dimensions of summary quality, which represents the average score given by three experts. Each data point also includes 11 reference summaries: the original summary from the CNN/DM dataset and 10 human-written abstractive summaries from \citet{kryscinski-etal-2020-evaluating}. 
For evaluation, we use system-level correlation using Kendall tau, where we first take the average score for each system and take the correlation.\footnote{
While they claimed to report system-level correlation, the BARTScore paper  \cite{yuan2021bartscore} actually calculated \textit{summary}-level correlation \cite{louis2013automatically}, where they first get correlation for each system and then take the average. Since we use evaluation metrics to rank \textit{systems}, we report system-level correlation following \citet{fabbri-etal-2021-summeval}.
}

\paragraph{Implementation Details}

The sentence matching functions are implemented as follows. We used the \texttt{rouge-score}\footnote{\url{https://pypi.org/project/rouge-score/}} Python library with default settings (i.e., without stemmer and no sentence-splitting of summaries with newlines) to calculate token-level ROUGE. 
We used the implementation of BLEU and CHRF in \texttt{sacrebleu}\footnote{\url{https://pypi.org/project/sacrebleu/}}.
We used the widely used default version of BERTScore\footnote{\url{https://github.com/Tiiiger/bert_score}}, which uses the \texttt{roberta-large} model in the \texttt{transformers} library \cite{wolf-etal-2020-transformers}. For T5-ANLI, we used the same implementation as in \citet{honovich2022true}, where T5-11B is fine-tuned with 25K training steps on ANLI \cite{nie-etal-2020-adversarial}, treating both contradiction and neutral pairs as not entailed. Finally, for BLEURT \cite{sellam-etal-2020-bleurt}, we used the \texttt{BLEURT-20} checkpoint\footnote{\url{https://github.com/google-research/bleurt}} suggested by the authors which also supports non-English languages.
Sentences are split using \texttt{nltk}\footnote{\url{https://pypi.org/project/nltk/}}.
For all experiments, we report f-measure scores whenever available, such as in ROUGE, BERTScore, and SMART. We also report the version of SMART that considers both source and reference as in Eq~\ref{eq:smart_with_src_ref}.

% \section{Results}

\begin{table}[th!]
    \small
    \centering
    \begin{tabular}{@{}l@{~~~}c@{~~~~}c@{~~~~}c@{~~~~}c@{~~~~}c@{~~~~}c@{}}
        \toprule
        Match Fn. & S. Type & Coh & Fac & Flu & Inf & $\mu$ \\
        \midrule
        \multicolumn{7}{c}{\textit{SMART with String-based Matching Functions}} \\
        \midrule
        ROUGE-1 & S1 & .233 & \textbf{.733}${}^*$ & .494 & {.500}${}^*$ & .490 \\
         & S2 & .217 & .683 & .477 & .417 & .448 \\
         & SL & .267 & {.700} & .460 & .467 & .473 \\
        ROUGE-2 & S1 & .183 & .650 & .477 & .417 & .432 \\
         & S2  & .183 & .650 & .477 & .417 & .432\\
         & SL & .217 & .650 & .444 & .383 & .423 \\
        ROUGE-L & S1 & .233 & \textbf{.733}${}^*$ & {.527}${}^*$ & {.500}${}^*$ & .498\\
         & S2 & .217 & .683 & .477 & .417 & .448 \\
         & SL & .267 & .700 & .460 & .467 & .473 \\
        BLEU & S1 & .300 & \textbf{.733}${}^*$ & {.527}${}^*$ & .467 & \underline{.507} \\
         & S2 & .283 & .717 & .510 & .450 & .490 \\
         & SL & .317 & .717 & .460 & .483 & .494 \\
        CHRF & S1 & .300 & \textbf{.733}${}^*$ & .494 & {.500}${}^*$ & \underline{.507} \\
         & S2 & .300 & .700 & .460 & .433 & .473 \\
         & SL & {.367}${}^*$ & \textbf{.733}${}^*$ & .494 & {.500}${}^*$ & \underline{.523} \\
        \midrule
        \multicolumn{7}{c}{\textit{SMART with Model-based Matching Functions}} \\
        \midrule
        BERTScore & S1  & .317 & .683 & .561 & .517 & .519  \\
         & S2 & -.017 & .383 & .276 & .183 & .207\\
         & SL & .383 & .683 & .527 & .583 & .544\\
        T5-ANLI & S1 & .117 & .550 & .444 & .350 & .365 \\
         & S2 & .133 & .533 & .360 & .333 & .340 \\
         & SL & .117 & .483 & .343 & .350 & 323 \\
        BLEURT & S1 & .433 & .667 & \textbf{.644}${}^*$ & .667 & \underline{.603} \\
         & S2 & .417 & \textbf{.750}${}^*$ & \textbf{.628}${}^*$ & .528 & \underline{.594} \\
         & SL & \textbf{.567}${}^*$ & .567 & .611 & \textbf{.733}${}^*$ & \underline{.619} \\
        \bottomrule        
    \end{tabular}
    \caption{Kendall tau system-level correlation of variants of SMART using different matching functions.
    For each block, correlations of metrics not significantly outperformed by any other metric (using William's Test; \citealp{graham-baldwin-2014-testing}) for that specific dimension are marked with an asterisk (*).
    Those that are not significantly outperformed by all metrics are \textbf{boldfaced}.
    We also show the average scores in the $\mu$ column, where the top three values for each block are \underline{underlined}.}
    \label{tab:matching_fn_results}
\end{table}

\subsection{SMART with Different Matching Functions}

We first compare different variants of SMART using the six matching functions described in Section \ref{sec:matching_fn}. Table \ref{tab:matching_fn_results} shows their system-level correlations, where Coh, Fac, Flu, and Inf stand for coherence, factuality, fluency, and informativeness, respectively. Among string-based matching functions, CHRF performs the best in terms of average correlation, followed by BLEU. This shows that machine translation metrics are better sentence matchers. Among model-based matching functions, BLEURT performs the best by a large margin; SMART with BLEURT significantly outperforms all the other model-based variants on all dimensions of summary quality. 
We believe that this is because BLEURT is optimized to match sentences, as well as to predict ROUGE, BLEU, BERTScore, and entailment scores.
Interestingly, T5-ANLI as a matching function underperforms even in the factuality dimension. We posit that this is because sentences are passed to the matching function without their neighboring context. %\sn{Without context? We are using both source and reference? Maybe because we use both source and reference, it is not working well with factuality.}

% ; an information 

\begin{table}[t!]
\small
\centering
\begin{tabular}{@{}lccccc@{}}
    \toprule
    Metric & {Coh} & {Fac} & {Flu} & {Inf} & $\mu$ \\
    \midrule
    \multicolumn{6}{c}{\textit{String-based Metrics}} \\
    \midrule
    ROUGE-1 & {.350} & .550 & {.527} & {.583} & .503 \\
    ROUGE-2 & .233 & .600 & .494 & .433 & .440 \\
    ROUGE-L & .117 & .117 & .259 & .350 & .211 \\
    BLEU & .217 & .050 & .326 & .383 & .244 \\
    CHRF & {.350} & {.617} & {.561} & {.550} & {.519} \\
    S1-CHRF & .300 & {\textbf{.733}} & {.494} & {.500} & {.507} \\
    S2-CHRF & .300 & {.700} & .460 & .433 & .473 \\
    SL-CHRF & {.367} & {\textbf{.733}} & {.494} & {.500} & \underline{.523} \\
    \midrule
    \multicolumn{6}{c}{\textit{Source-free Model-based Metrics}} \\
    \midrule
    BERTScore & .333 & -.030 & .142 & .200 & .161 \\
    MoverScore & .217 & -.050 & .259 & .350 & .194 \\
    BLEURT & \textbf{.533} & .200 & {.410} & {.467} & .403 \\
    SMS & .267 & {.600} & .360 & .400 & \underline{.407} \\
    \midrule
    \multicolumn{6}{c}{\textit{Source-dependent Model-based Metrics}} \\
    \midrule
    PRISM & .233 & .600 & .360 & .367 & .390 \\
    Q${}^2$ & .250 & \textbf{.750} & .577 & .450 & .507 \\
    T5-ANLI & .250 & .583 & .544 & .517 & .473 \\
    BARTScore & .350 & {.617} & .494 & .450 & .478 \\
    BARTScore+CNN${}^\dagger$ & \textbf{.550} & .317 & {.594} & {.583} & {.511} \\
    S1-BLEURT & .433 & {.667} & \textbf{.644} & {{.667}} & {{.603}} \\
    S2-BLEURT & .417 & \textbf{.750} & {\textbf{.628}} & {.583} & {.594} \\
    SL-BLEURT & \textbf{.567} & .567 & {.611} & \textbf{.733} & \underline{.619} \\
    \bottomrule
\end{tabular}
\caption{Kendall tau system-level correlation of different metrics on the SummEval dataset.
Correlations of metrics not significantly outperformed by any other metric (using William's Test; \citealp{graham-baldwin-2014-testing}) for that specific dimension are \textbf{boldfaced}.
We also show the average scores in the $\mu$ column, where the best values for each block are \underline{underlined}.
Note that BARTScore+CNN uses BART that is fine-tuned on CNN/DM, the same data source as SummEval, thus direct comparison with the other metrics is not possible.}
\label{tab:main_results}
\end{table}

\subsection{SMART Compared to Other Text Evaluation Metrics}

We compared SMART with three types of metrics: string-based metrics, source-free and source-dependent model-based metrics.

{\em String-based metrics} include: (1-3) ROUGE-1/2/L \cite{lin-2004-rouge} measuring token-level overlap between reference and output summaries; 
% : Token-level ROUGE metrics that measures unigram and bigram overlap, as well as the longest common subsequence of tokens; 
(4) BLEU \cite{papineni-etal-2002-bleu} measuring token-level overlap between reference and output summaries with a focus on precision; 
% : calculates token-based n-gram overlap with a focus on precision;
(5) CHRF \cite{popovic-2015-chrf} measuring character-based n-gram overlap  between reference and output summaries; and (6-8) S1/2/L-CHRF, our best SMART metric using a string-based matching function from Table~\ref{tab:matching_fn_results}. 

{\em Source-free model-based metrics} include: (9) BERTScore \cite{Zhang*2020BERTScore:}: A metric that relies on BERT(-like) models \cite{devlin-etal-2019-bert} and computes an aggregation of the token-level similarity scores; (10) MoverScore \cite{zhao-etal-2019-moverscore}: measures the semantic distance between BERT n-gram embeddings of reference and candidate summaries using Word Mover's Distance (WMD; \citealp{kusner2015from}); (11) BLEURT \cite{sellam-etal-2020-bleurt}: finetunes BERT using a combination of real and synthetic training data with gold-/silver-standard human judgment scores; and (12) Sentence Mover's Similarity (SMS; \citealp{clark-etal-2019-sentence}): uses an extension of WMD that works with sentences instead of tokens.

{\em Source-dependent model-based metrics} include: (13) PRISM \cite{thompson-post-2020-automatic}: leverages a zero-shot paraphrasing model and uses probabilities from force-decoding the candidate summary given the source as input;
(14) Q${}^2$ \cite{honovich-etal-2021-q2}: employs question generation and question answering models and checks whether answers from the summary are entailed by answers from source;
(15) T5-ANLI \cite{honovich2022true}: fine-tunes T5 \cite{raffel2020exploring} using the ANLI dataset \cite{nie-etal-2020-adversarial} to produce an entailment score given the source as premise and the summary as hypothesis;
(16-17) BARTScore(+CNN) \cite{yuan2021bartscore}: evaluates text using probabilities from force-decoding the candidate summary given the source as input using BART \cite{lewis-etal-2020-bart} without (with) finetuning with CNN/DM summarization dataset \cite{hermann2015teaching}; and (18-20) S1/2/L-BLEURT: Our best SMART metric using a model-based matching function from Table~\ref{tab:matching_fn_results}. 

Table \ref{tab:main_results} reports the system-level correlations of different metrics for each quality dimension. For all quality dimensions, SMART with BLEURT matching function has the highest correlation, where SL-BLEURT evaluates coherence and informativeness better, and S1-BLEURT and S2-BLEURT evaluate factuality and fluency better, respectively. On average, SL-BLEURT performs best, followed by S1-BLEURT and S2-BLEURT, all three of which outperforming BARTScore+CNN, which is finetuned on the same summarization dataset as SummEval.
S2-BLEURT also performs comparably with Q${}^2$ in factuality evaluation.
Given that each of the SMART metrics are better at evaluating different quality dimensions, it is therefore recommended to use them as a set, similar to how ROUGE metrics are used. 
We can also see in the table that source-dependent metrics are better than source-free ones, signifying the importance of the use of source documents during summary evaluation.
Among source-free metrics, SMS performs the best, showing the superiority of sentence-level matching metrics against the token-level ones. Among string-based metrics, SL-CHRF performs the best, with a competitive system-level correlation when compared with BARTScore+CNN on average. This shows that SMART achieves comparable evaluation power with previous LM-based metrics even without using any pretrained language models.

\subsection{Ablation Studies}
\label{sec:ablation}

We present various ablation studies on the different components of SMART in Table \ref{tab:ablation}. For simplicity, we report on SMART-\texttt{X}, an average of SMART-l, SMART-2 and SMART-L.\footnote{From here on, we use $\texttt{X}$ to correspond to the average of the three \texttt{[1|2|L]} variants of ROUGE or SMART.}
The first block contains SMART-\texttt{X} that uses only either precision or recall, both of which have substantially lower system-level correlation than f-measure. The recall variant performs better, which follows results from traditionally recall-oriented evaluation metrics in summarization \cite{lin-2004-rouge}. 

\begin{table}[t]
\small
\centering
\begin{tabular}{@{}l@{~~~}c@{~~~~}c@{~~~~}c@{~~~~}c@{~~~~}c@{}}
    \toprule
    Component & Coh & Fac & Flu & Inf & $\mu$\\
    \midrule
    S\texttt{X}-BLEURT & {.450} & {.750} & {.661} & {.650} & {.628} \\
    \midrule
    \multicolumn{6}{c}{\textit{using {precision} or recall}} \\
    \midrule
    precision & .083 & .017 & .259 & .350 & .177 \\
    recall & .283 & .683 & .410 & .383 & .440 \\
    \midrule
    \multicolumn{6}{c}{\textit{using a different source/reference aggregation}} \\
    \midrule
    ref-only & -.100 & -.300 & -.092 & -.033 & -.131 \\
    src-only & .267 & .700 & .460 & .433 & .465 \\
    average & .400 & {.750} & .577 & .600 & .582 \\
    minimum & .267 & .700 & .427 & .467 & .465  \\
    \midrule
    \multicolumn{6}{c}{\textit{using a different reference summary}} \\
    \midrule
    ref-only (CNN/DM) & -.100 & -.300 & -.059 & -.033 & -.123 \\
    ref-only (best system) & .367 & .100 & .410 & .500 & .344 \\
    max (CNN/DM) & .500 & .767 & .711 & .667 & .661 \\
    max (best system) & .667 & .467 & .745 & .867 & .686 \\
    \bottomrule
\end{tabular}
\caption[Caption for LOF]{Ablation study. Kendall tau system-level correlation of S\texttt{X}-BLEURT when one component is set to a different configuration. } 
\label{tab:ablation}
\end{table}

The second block contains different ways to aggregate SMART scores that compare candidate summaries (1) to the source documents and (2) to the reference summaries. When using only one of the two scores (i.e., reference- or source-only), SMART scores significantly perform worse, which implies that using both source and reference is necessary for SMART to work. We also tried aggregation through taking the average or the minimum, both of which perform worse than taking the maximum as in Eq~\ref{eq:smart_with_src_ref}.

One interesting finding is that using only the reference summaries to calculate SMART gives negative correlation scores on all dimensions. 
Given the findings from \citet{fabbri-etal-2021-summeval} that CNN/DM summaries are worse than system summaries in terms of human ratings, we further investigated using different reference summaries for SMART.
Specifically, we tried (1) replacing the set of reference summaries from SummEval into the original summary from CNN/DM, and (2) treating the best system summary according to the average human score as the reference. As can be seen in the third block of Table \ref{tab:ablation}, CNN/DM reference summaries correlate negatively on all dimensions.
When using the best system summary as the reference, SMART now obtains positive correlations across all dimensions. The correlation further improves when SMART uses both the source and the new reference.

\section{Further Analyses}
\label{sec:analysis}

% In the following subsections, we further analyse how SMART works well with longer documents and summaries, is less biased towards specific models, and selects better model checkpoints than ROUG

% we compare the following metrics: ROUGE-$\texttt{X}$, the widely used summarization metric, S$\texttt{X}$-CHRF, SMS, and BARTScore, top performing string-based metrics, and source-free and source-dependent model-based metrics, and S$\texttt{X}$-BLEURT, our best SMART metric.

\begin{figure*}[t!]
    \centering
    \includegraphics[width=\textwidth]{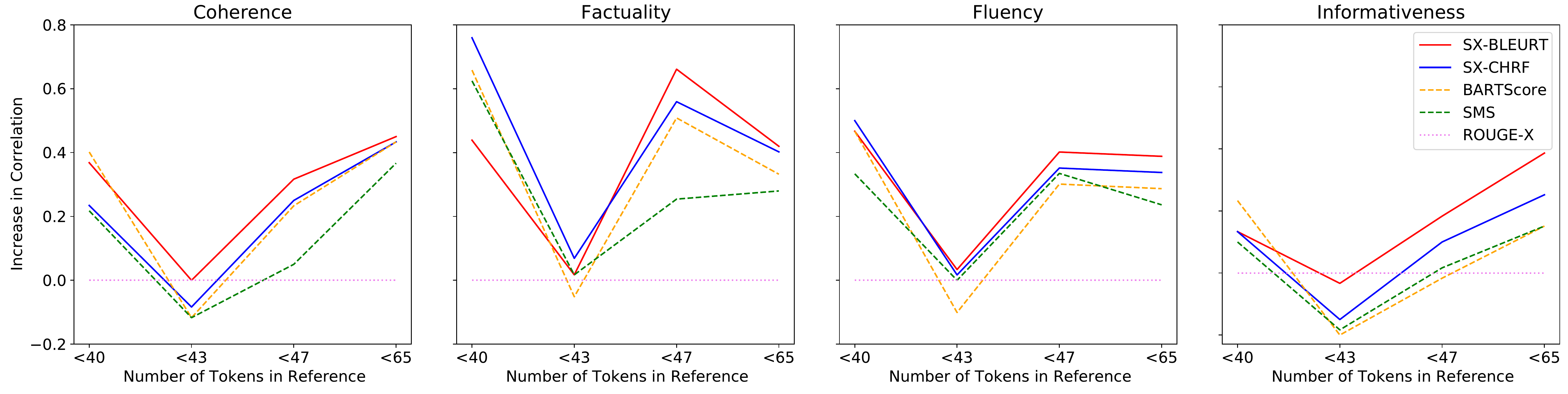}
    \caption{Length Analysis. Relative increase in 
    Kendall tau system-level correlation with respect to ROUGE-$\texttt{X}$ of four evaluation metrics for each length bucket (leftmost bucket has the shortest lengths).}
    \label{fig:length_analysis}
\end{figure*}

\paragraph{SMART Works Well with Longer Summaries}
% The first experiment investigates the performance of the metrics with respect to the length of both the source and the reference. 
We divided the datasets into four buckets based on the average number of tokens in the reference summary, where the first bucket contains the shortest reference summaries. For each bucket, we then calculated system-level correlation for all competing metrics. 
For each quality dimension, we report the relative increase in correlation with respect to ROUGE-$\texttt{X}$, which is illustrated in Figure \ref{fig:length_analysis}. As can be seen in the figure, in general, all metrics perform better relative to ROUGE as the number of tokens increases from 43 tokens, which shows that ROUGE is not suitable for long summary evaluation. 
Interestingly, ROUGE also underperforms in the first bucket, which means that it is also not good at evaluating short summaries.
Among the competing metrics, S$\texttt{X}$-BLEURT (and S$\texttt{X}$-CHRF) correlate the best (and second best) when the there are more tokens in the source/reference. %\sn{(Drop source, not studied yet. And just say that We have not studied this yet, but input lengths will not be a concern and the metric will be a good fit for evaluating long-input tasks.)}.

\begin{figure*}[t!]
    \centering
    \includegraphics[width=\textwidth]{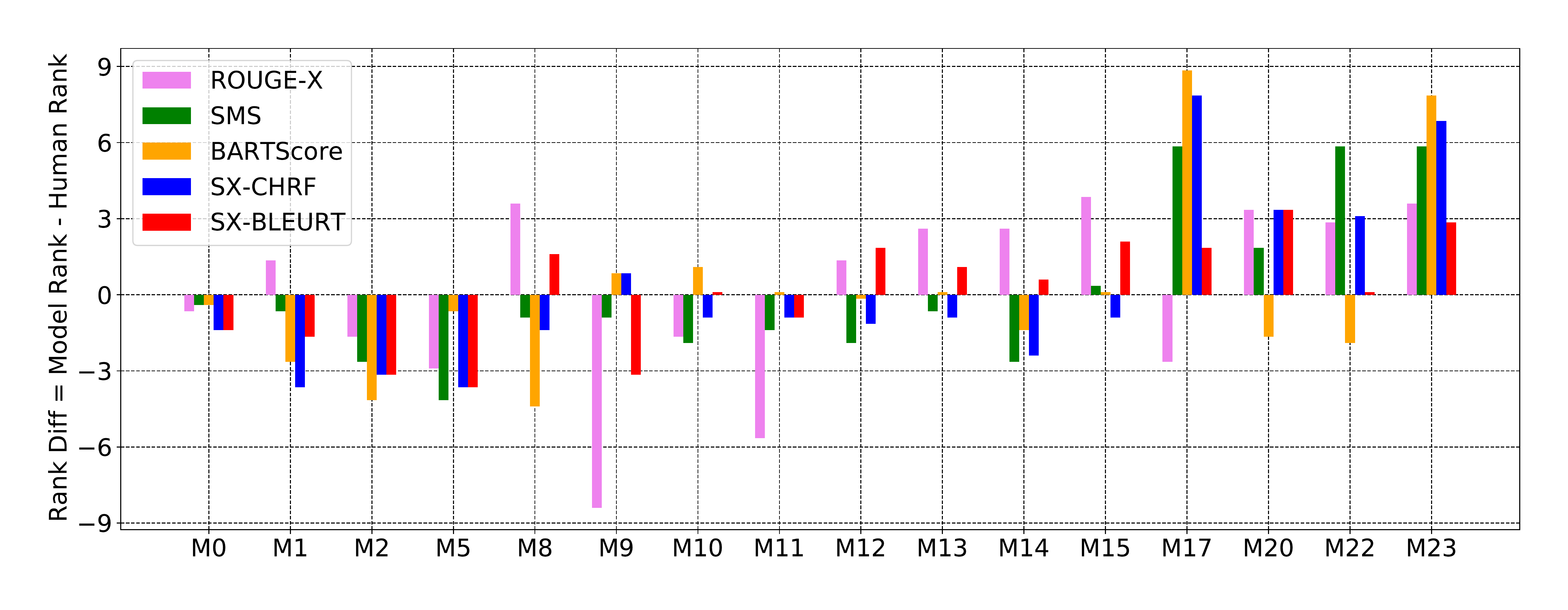}
    \caption{Bias Analysis. Difference in ranking between human scores and metric scores for each competing metric, averaged over all quality dimensions. A negative value means that the metric ranks the system higher. M0 to M23 in the x-axis correspond to the summarization models, where M0 to M5 are extractive and the rest are abstractive. See \citet{fabbri-etal-2021-summeval} for detailed model descriptions.}
    \label{fig:bias_analysis}
\end{figure*}

\begin{table}[t!]
    \small
    \centering
    \begin{tabular}{@{}clcc@{}}
        \toprule
        & Metric & $\sigma$(Rank Diff) $\downarrow$ & Pairwise Acc. $\uparrow$ \\
        \midrule
        & ROUGE-$\texttt{X}$ & 5.012 & 63.33 \\
        & SMS & 5.500 & 63.33 \\
        & BARTScore & 4.717 & 67.50 \\
        & S$\texttt{X}$-CHRF & 5.232 & 65.00 \\
        \raisebox{4ex}[0pt]{\begin{sideways}Coh\end{sideways}} & S$\texttt{X}$-BLEURT & \textbf{4.228} & \textbf{72.50} \\
        \midrule
        & ROUGE-$\texttt{X}$ & 4.228 & 70.00 \\
        & SMS & 2.915 & 80.83 \\
        & BARTScore & 2.915 & 80.83 \\
        & S$\texttt{X}$-CHRF & \textbf{2.062} & 86.67 \\
        \raisebox{4ex}[0pt]{\begin{sideways}Fac\end{sideways}} & S$\texttt{X}$-BLEURT & {2.151} & \textbf{87.50} \\
        \midrule
        & ROUGE-$\texttt{X}$ & 3.969 & 72.50 \\
        & SMS & 4.650 & 67.50 \\
        & BARTScore & 4.031 & 75.00 \\
        & S$\texttt{X}$-CHRF & 3.824 & 74.17 \\
        \raisebox{4ex}[0pt]{\begin{sideways}Flu\end{sideways}} & S$\texttt{X}$-BLEURT & \textbf{2.646} & \textbf{83.33} \\
        \midrule
        & ROUGE-$\texttt{X}$ & 3.841 & 75.00 \\
        & SMS & 4.783 & 70.00 \\
        & BARTScore & 4.500 & 72.50 \\
        & S$\texttt{X}$-CHRF & 4.198 & 75.00 \\
        \raisebox{4ex}[0pt]{\begin{sideways}Inf\end{sideways}} & S$\texttt{X}$-BLEURT & \textbf{2.739} & \textbf{82.50} \\
        \bottomrule
    \end{tabular}
    \caption{The standard deviation of the difference in ranking between human and metric scores $\sigma$(Rank Diff) and the pairwise ranking accuracy of different metrics for different summary quality. Best scores are \textbf{boldfaced}.}
    \label{tab:bias_analysis}
\end{table}

\paragraph{SMART is Less Biased towards Specific Models}
% In this section, we examine whether the competing metrics are biased towards certain types of summarization models. 
While we acknowledge that all automatic metrics are not perfect as shown in Table \ref{tab:main_results}, their rankings should not be hugely different from human rankings. Moreover, they should not be biased towards a single summarization model. To this end, we get the difference in rankings given by humans and by the automatic metrics for each summarization model for each quality dimension. Figure \ref{fig:bias_analysis} illustrates the resulting differences averaged over all dimensions of summary quality, in which we have two interesting observations. Firstly, all automatic metrics are in general biased towards ranking extractive systems higher. This suggests that a separate extractive and abstractive model evaluation is necessary using current automatic metrics. We leave exploration of metrics that are equally unbiased towards both extractive and abstractive for future work.
Secondly, we found that BARTScore scores BART (M22 in Figure~\ref{fig:bias_analysis}; \citealp{lewis-etal-2020-bart}) significantly higher than all the other models (BARTScore of BART is $-1.398$ vs. $-2.303 \pm 0.313$ on average without BART). This problem is amplified when BARTScore is finetuned using the CNN/DM dataset ($-0.488$ vs. $-1.702 \pm 0.248$). This shows that using pretrained encoder-decoder models for summary evaluation induces bias towards summarization models finetuned on the same model.

In Figure \ref{fig:bias_analysis}, we can see that S$\texttt{X}$-BLEURT is the least biased since its rank differences with human scores are closer to zero.
To quantitatively measure if the above statement is true, we use two measures. The first one is the standard deviation of the rank difference, where the score closest to zero can be considered the least biased. The second measure is the pairwise ranking accuracy, where for all pairs of system, we check whether human and metric rankings are equivalent. Table \ref{tab:bias_analysis} shows these numbers, which show that S$\texttt{X}$-BLEURT has the lowest standard deviation of the rank difference and the highest pairwise rank accuracy across all quality dimensions. This entails that the metric is the least biased among the competing metrics.

\section{Conclusions}

In this paper, we proposed SMART (Sentence Matching for Rating Text), a new metric for evaluating generated text given both a source document and a reference text. SMART makes use of a sentence-level soft-matching function to match sentences, which can either be string-based or model-based depending on the available resources. This function can easily be replaced with new and better ones, which allows us to create a better SMART metric. We provided two types of SMART based on n-gram overlap and longest common subsequence, and our extensive experiments showed that SMART evaluates document summaries better in terms of all four dimensions of summary quality: coherence, factuality, fluency, and informativeness. Our analyses also showed that SMART is better as summary length increases and is less biased than other competing metrics.
%and selects better checkpoints than ROUGE. 
Based on our findings, we recommend using SMART-\texttt{[1|2|L]}-BLEURT for high quality evaluation (e.g., when comparing the final model with the state of the art), and using SMART-\texttt{[1|2|L]}-CHRF for fast evaluation (e.g., ablation studies and model development phases). Both versions are also suitable for multilingual texts, since BLEURT was finetuned using a multilingual BERT and CHRF is designed to handle morphologically-rich languages.

\subsection{Limitations and Future Work}

We acknowledge several limitations of our work. Firstly, while our proposed metric is designed to work on text generation tasks including long-form question answering and dialog response generation, our experiments are limited to using a news summarization meta-evaluation suite. This is due to the availability of meta-evaluation suite that holistically evaluates text on different dimensions of text quality. We plan to create a meta-evaluation suite using datasets with longer outputs and from non-news domains.

Currently, SMART assumes single-source inputs, which makes it non-trivial to use for evaluating texts in tasks with multi-source inputs. While this can be mitigated by concatenating documents in chronological order, there may be cases where such ordering is not possible. Ultimately, finding an efficient way to extend the (soft version of) longest common subsequence algorithm into multiple sources is necessary, which we plan to explore in the future.

Finally, we plan to maintain an open-source toolkit which includes scripts to easily run evaluation using SMART, and is updated accordingly when new features, such as better matching functions or support for multi-source inputs.

% 

% future work
%% meta-evaluation suite for longer summaries and for non-news domain
%% a way to handle multiple documents
%% optimizing models with Sent-ROUGE
%\input{350limitations}

\bibliography{anthology,custom}
\bibliographystyle{acl_natbib}

\end{document}